\documentclass[letterpaper]{article} 
\usepackage{aaai24}  
\usepackage{times}  
\usepackage{helvet}  
\usepackage{courier}  
\usepackage[hyphens]{url}  
\usepackage{graphicx} 
\usepackage{natbib}  
\usepackage{caption} 
\usepackage{algorithm}
\usepackage{algorithmic}

\usepackage{newfloat}
\usepackage{listings}
\DeclareCaptionStyle{ruled}{labelfont=normalfont,labelsep=colon,strut=off} 
\floatstyle{ruled}
\newfloat{listing}{tb}{lst}{}
\floatname{listing}{Listing}
\usepackage{epsfig}
\usepackage{multirow}
\usepackage{amssymb}
\usepackage{enumitem}

\usepackage{booktabs} 

\usepackage[table, svgnames, dvipsnames]{xcolor}
\usepackage{amsmath}
\usepackage{pifont}
\definecolor{darkergreen}{RGB}{19, 182, 53}
\newcommand\gain[1]{{\textcolor{darkergreen}{(#1{$\uparrow$})}}}

\usepackage{hhline}
\usepackage{capt-of,etoolbox}

\title{DiffusionEngine: Diffusion Model is Scalable Data Engine for Object Detection\\ \textcolor{magenta}{\Large Project Page: \textit{mettyz.github.io/DiffusionEngine}}}
\author{
    Manlin Zhang\textsuperscript{\rm 1,2}\equalcontrib,
    Jie Wu\textsuperscript{\rm 2}\equalcontrib$^\dagger$,
    Yuxi Ren\textsuperscript{\rm 2}\equalcontrib,
    Ming Li\textsuperscript{\rm 2},
    Jie Qin\textsuperscript{\rm 2},
    Xuefeng Xiao\textsuperscript{\rm 2},\\ \vspace{2pt}
    Wei Liu\textsuperscript{\rm 2},
    Rui Wang\textsuperscript{\rm 2},
    Min Zheng\textsuperscript{\rm 2},
    Andy J. Ma\textsuperscript{\rm 1}$^\dagger$
} 
\affiliations{
    \textsuperscript{\rm 1}Sun Yat-sen University\hspace{2pt}
    \textsuperscript{\rm 2}ByteDance Inc\\
}

\begin{document}

\twocolumn[{%
\renewcommand\twocolumn[2][]{#1}%
\maketitle%
\vspace*{-1cm}
    \centering\centering
    \includegraphics[width=0.98\textwidth,height=9.8cm]{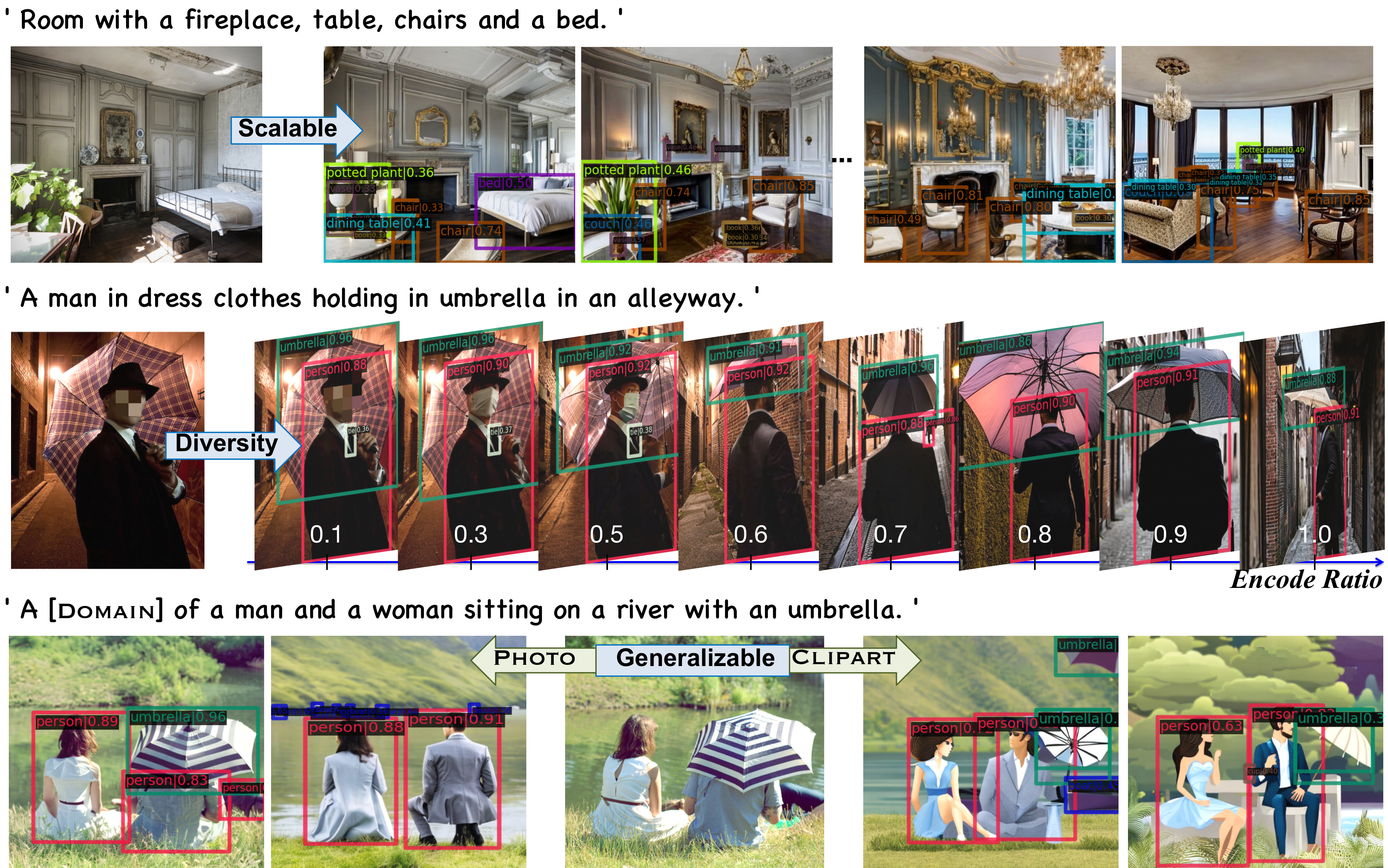}
    \vspace{0.06cm}
    \captionof{figure}{
We propose \textbf{DiffusionEngine} to scale up high-quality detection-oriented training pairs. DiffusionEngine is \textit{scalable} (1st row), \textit{diverse} (2nd row), and can \textit{generalize} robustly across domains (3rd row).
}
\vspace*{0.5cm} 
\label{fig:head}
}]%

\renewcommand{\thefootnote}{}
\footnotetext{$^\star$Equal contribution. $^\dagger$Corresponding author.
}

\begin{abstract}
\vspace*{-.1cm}
Data is the cornerstone of deep learning.
This paper reveals that the recently-developed Diffusion Model is a scalable data engine for object detection. 

Existing methods for scaling up detection-oriented data often require manual collection or generative models to obtain target images, followed by data augmentation and labeling to produce training pairs, which are costly, complex, or lacking diversity.
To address these issues, we present \textbf{DiffusionEngine} (DE), a data scaling-up engine that provides high-quality detection-oriented training pairs in a single stage.
DE consists of a pre-trained diffusion model and an effective \textbf{Detection-Adapter}, contributing to generating scalable, diverse and generalizable detection data in a plug-and-play manner.
Detection-Adapter is learned to align the implicit semantic and location knowledge in off-the-shelf diffusion models with detection-aware signals to make better bounding-box predictions. 
Additionally, we contribute two datasets, \emph{i.e.}, \textbf{COCO-DE} and \textbf{VOC-DE}, to scale up existing detection benchmarks for facilitating follow-up research.
Extensive experiments demonstrate that data scaling-up via DE can achieve significant improvements in diverse scenarios, such as various detection algorithms, self-supervised pre-training, data-sparse, label-scarce, cross-domain, and semi-supervised learning.
For example, when using DE with a DINO-based adapter to scaling-up data, mAP is improved by \textbf{3.1\%} on COCO, \textbf{7.6\%} on VOC and \textbf{11.5\%} on Clipart.
\end{abstract}
\vspace{-0.5cm}
\section{Introduction}
Recent years have witnessed the prevalence of object detection in extensive vision applications such as scene recognition and understanding.
However, the success of these applications based on object detection heavily relies on high-quality training data of images with granular box-level annotations. 
The traditional practice for obtaining such data involves manual annotations for a massive number of images collected from the web, which is expensive, time-consuming, and expert-involved. 
Furthermore, the images from real-world scenarios often follow a data-sparse, long-tail, or out-of-domain distribution, raising more uncertainty and difficulty in this traditional data collection paradigm.

Recently, the diffusion model has shown great potential in image generation and stylization, and researchers have explored its use in assisting object detection tasks.
For example, DALL-E for detection~\cite{ge2022dall}  generates the foreground objects and the background context separately, and then employs copy-paste technology to obtain synthetic images.
Similarly, X-Paste~\cite{zhao2022x} copies generated foreground objects and pastes them into existing images for data expansion.
However, these existing solutions have several drawbacks:
i) Additional expert models are required for labeling, increasing the complexity and cost of the data scaling process.
ii) These methods naively paste the generated objects into repeated images, resulting in limited diversity and producing unreasonable images.
iii) Image and annotation generation processes are separated, without fully leveraging the detection-aware concepts of semantics and location learned from the diffusion model.
These issues prompt us to raise the question:
\textit{how to design a more straightforward, scalable, and effective algorithm for scaling up detection data?}

To address this issue, we propose a novel tool called DiffusionEngine, comprising a pre-trained diffusion model and a Detection-Adapter.
We reveal that the pre-trained diffusion model has implicitly learned object-level structure and location-aware semantics, which can be explicitly utilized as the backbone of the object detection task. 
Furthermore, the Detection-Adapter can be constructed through diverse detection frameworks, enabling the acquisition of detection-oriented concepts from the frozen diffusion-based backbone to produce precise annotations.
Our contributions are summarized as follows:

\begin{itemize}[ itemsep = 0.6pt]
\vspace{-0.1cm}
\item ~\underline{\textit{New Insight}}: We propose DiffusionEngine, a simple yet effective engine for scaling up object detection data. By abandoning complex multi-stage processes and instead designing a Detection-Adapter to generate training pairs in a single stage, DiffusionEngine is both efficient and versatile. Moreover, it is orthogonal to most detection works and can be used to improve performance further in a plug-and-play manner.

\vspace{-0.05cm}
\item ~\underline{\textit{Pioneering and Scalable}}: Detection-Adapter aligns the implicit knowledge learned by off-the-shelf diffusion models with task-aware signals, empowering DiffusionEngine with excellent labeling ability. Furthermore, DiffusionEngine has an infinite capacity for scaling up data, with the ability to expand tens of thousands of data.

\vspace{-0.05cm}
\item ~\underline{\textit{Novel Dataset}}: To facilitate further research on object detection, we contribute two scaling-up datasets using DiffusionEngine, namely COCO-DE, and VOC-DE. These datasets scale up the original images and annotations, which provides scalable and diverse data for leading-edge research  to enable the next generation of state-of-the-art detection algorithms.

\item ~\underline{\textit{High Effectiveness}}: Experiments demonstrate that DiffusionEngine is scalable, diversified, and generalizable, achieving significant performance improvements under various settings. We also reveal that DiffusionEngine is superior to traditional methods, multi-step approaches, and Grounding Diffusion Models in data scaling up.

\end{itemize}

\begin{figure}[t]
	\centering
	\includegraphics[width=1.\linewidth]{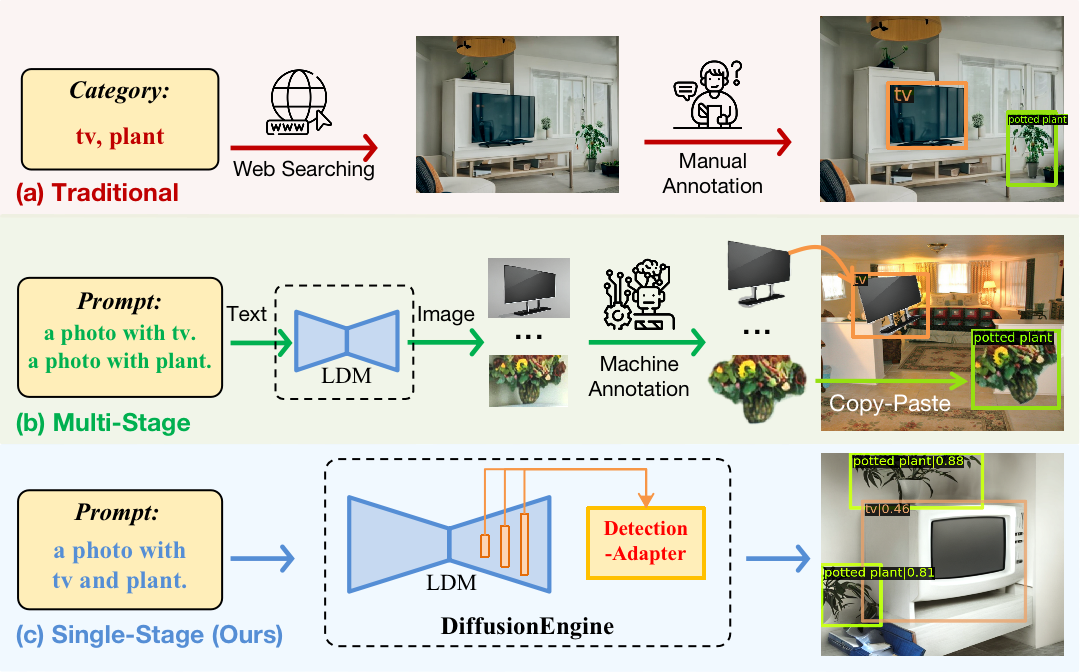}
	\caption{{Comparing the proposed DiffusionEngine with other data collection pipelines for object detection}.
	(a): Traditional pipeline is time-consuming and expert-involved. 
	(b): Existing multi-stage methods contain object-centric image generation, segmentation labeling, and copy-paste. It reduces human participation while introducing extra models and producing unconscionable images.
	(c): Our method generates reliable images and annotations in single stage. 
	}
 	\label{fig:teaser}
\end{figure}

\section{Related Works}
\subsection{Object Detection}
In recent years, the research area of object detection has seen significant advancements with the advent of convolutional neural networks~\cite{krizhevsky2017imagenet}.
Most object detection methods~\cite{girshick2015fast, ren2015faster, cai2019cascade, redmon2016you, liu2016ssd} can be broadly categorized into two paradigms: two-stage and one-stage methods. The two-stage models~\cite{girshick2015fast, cai2019cascade, ren2015faster, he2017mask} first generate box proposals and then perform regression and classification. 
In contrast, one-stage models~\cite{redmon2016you,lin2017focal,tian2019fcos,liu2016ssd} simultaneously predict the position and class probability of the detection boxes based on the priors of anchors or object centers. 
Furthermore, the emergence of Transformer~\cite{vaswani2017attention} leads to the development of transformer-based detection methods~\cite{carion2020end, zhu2020deformable}, which aim to define the detection task as a sequence prediction problem. 

\subsection{Scaling Up Data for Object Detection}
Large-scale high-quality training images and annotations are the keys to advanced detectors. However, real data distribution faces many challenges, such as few-shot and long-tailed. To alleviate this issue, 
various techniques for scaling up detection data are explored,
including data augmentation~\cite{wang2019data, zoph2020learning, chen2021scale} and data synthesis with generative models~\cite{ghiasi2021simple, zhao2022x, ge2022dall}. 
However, such methods generate new data using the original data as raw material, which leads to a lack of diversity. 
On the other hand, generative models can generate diverse  data that never appeared in the original dataset.
\cite{ge2022dall} leverages a powerful text-to-image generative model to generate diverse foreground objects and backgrounds, which are then composited to synthesize training data.

\subsection{Generative Models}
Generative models, such as generative adversarial networks (GAN)~\cite{creswell2018generative}, variational autoencoders (VAE)~\cite{kingma2013auto}, and flow-based models~\cite{kingma2018glow}, have seen significant advancements in recent years. 
Recently, Diffusion Probabilistic Models (DPM)~\cite{ho2020denoising, sohl-dickstein15} have emerged as a promising research direction, demonstrating their ability to generate high-quality images on diverse datasets~\cite{ho2022cascaded, nichol2021improved, saharia2022palette}. These methods are trained on billions of image-caption pairs for text-to-image generation tasks, such as DALL-E 2~\cite{ramesh2022hierarchical}, Imagen~\cite{saharia2022photorealistic}, and Stable Diffusion~\cite{rombach2022high}.
Although existing work~\cite{ge2022dall} has explored the use of DPM for detection-oriented training data synthesis, it separates the image and label generation stages.
This paper takes the first step to generate high-quality detection training pairs in a single stage.
\section{Method}
In the following subsections, we first present the preliminary of 
 latent diffusion model (LDM)~\cite{rombach2022high}.
Then, we reveal that LDM is an effective and robust backbone for object detection and details our novel strategy to effectively learn the Detection-Adapter by using existing detection datasets. 
At last, we present the way to use our DiffusionEngine for scaling up detection data.

\subsection{Preliminary}
\label{sec:preliminary}
We leverage the off-the-shelf pre-trained LDM to generate high-quality images in this work. 
LDM is a conditional image generator that includes an autoencoder for perceptual compression and a diffusion probabilistic model in the latent space.
It is built with the U-Net backbone modulated via a cross-attention mechanism for text-guided image generation. 
The process of image-guided text-to-image generation consists of four stages: i) encoding the image to the latent space $z_0=\mathcal{E}(x)$; ii) obtaining a noisy sample by forward diffusion; iii) getting a clean sample via backward diffusion; and iv) decoding the latent vector back to the image $x=\mathcal{D}(z_0)$. 
As shown in~\cite{ho2020denoising, lu2022dpm}, the forward diffusion has a closed-form solution to obtain the noise sample in any time step $t$,
\begin{equation}
\label{eq:forward_diffusion}
    z_t = \alpha_t z_0 + \sigma_t \epsilon, \quad\epsilon\sim\mathcal{N}(\epsilon|\mathbf{0}, \mathbf{I}),
\end{equation}
where $\alpha_t, \sigma_t\in\mathbb R^+$ are differentiable functions of $t$ with bounded derivatives, whose choice is determined by the noise schedules of the sampler.
In the backward diffusion process, a sequence of denoising steps is performed to progressively obtain cleaner samples, with the U-Net estimating the noise added at each time step. Each denoising step can be represented by the following function:
\begin{equation}
\label{eq:backward_diffusion}
    z_{t-1} = \epsilon_\theta(z_t, t, c_p),
\end{equation}
where $\epsilon_\theta$ refers to the U-Net, and $c_p=\tau_\theta(P)$ is the embedding of the input text prompt $P$. The U-Net blocks consist of a residual block, a self-attention block, and a cross-attention block. The text-condition $c_p$ is injected via each cross-attention as both the \textit{Key} and \textit{Value}, \emph{i.e.},
\begin{equation}
    Attention_i(Q_i, K_i, V_i) = Attention(\varphi_i(z_t), c_p, c_p),
\end{equation}
where $\varphi_i(z_t)$ is the visual representation in the $i^{th}$ U-Net block.
In the following paper, when referring to extracting intermediate features, we mean extracting the outputs of each U-Net block unless otherwise stated.

\begin{figure}[b]
	\centering
	\includegraphics[width=1.\linewidth]{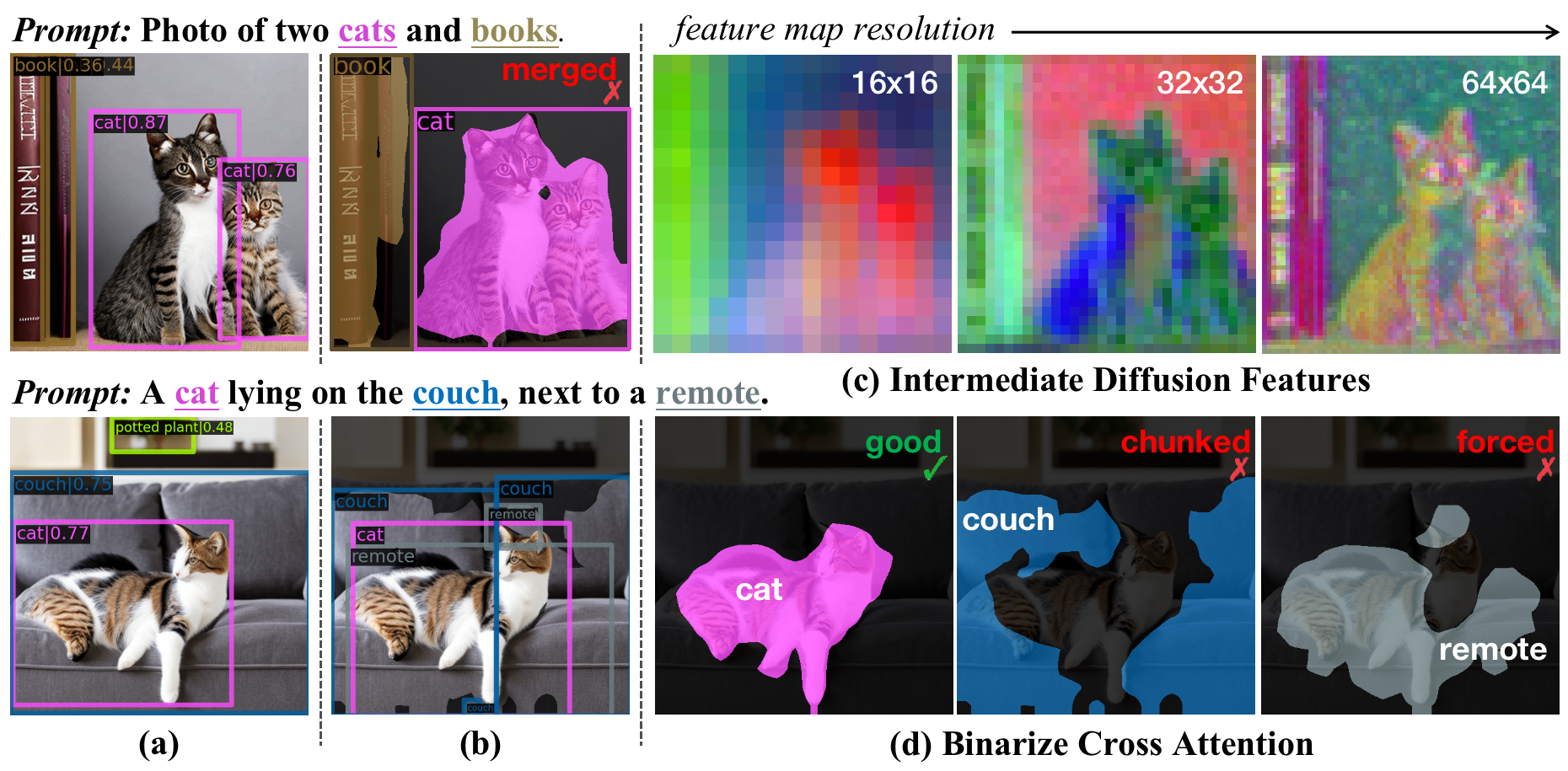}
	\caption{
(a)(b): Comparing our DE to cross-attention for detection, we find DE produces more accurate and less noisy results. (c): Different U-Net decoding stages capture coarse-to-fine object features. (d): Simply binarizing the cross-attention for each object highlights the semantic relative areas but is not sufficient for finer object detection.
}
 	\label{fig:feature}
\end{figure}
\subsection{LDM is Effective Backbone for Detection}
\label{sec:rational}

In this section, we analyze the \textit{location} and \textit{semantic} information contained in LDM and reveal that the pre-trained LDM has implicitly learned detection-oriented signals.

\noindent\textbf{The {location} information in LDM} is implicitly encoded in the LDM features.To illustrate this more effectively, we visualize the first three primary components of the extracted feature maps from different denoising stages in Fig.~\ref{fig:feature}(c).
At lower resolutions (\emph{e.g.}, 16x16), a coarse layout for objects in the image can be observed, with pixels in the same category sharing similar colors.
As the resolution increases, finer-grained location signals become more prominent, allowing for more precise object instance detection.

\begin{figure*}[th]
\centering
\includegraphics[width=.96\linewidth]{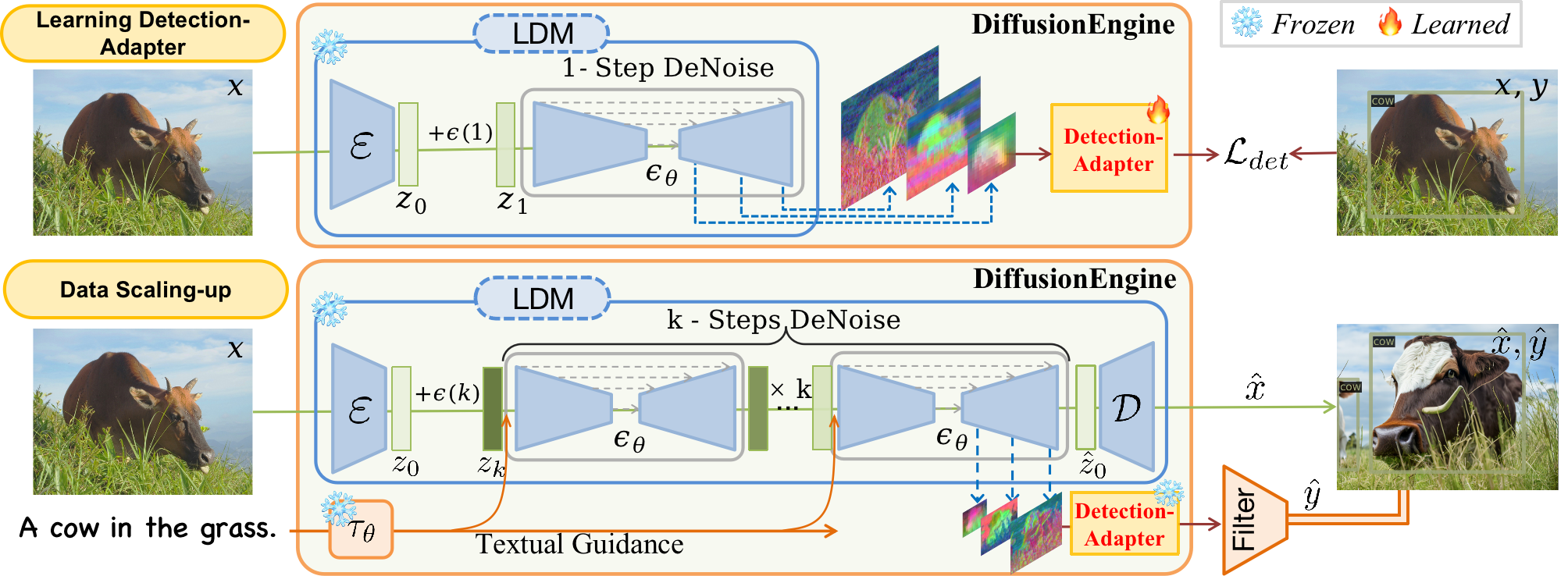}
\caption{Overview of the proposed DiffusionEngine. 
\textbf{The upper figure} shows the training procedure of DiffusionEngine. 
Each image undergoes a 1-step noise adding and then denoising to simulate the last image generation step in the LDM. The detection adapter learns to leverage the extracted pyramid features from the U-Net for detection. 
\textbf{The figure below} shows how we use the trained DiffusionEngine for data scaling-up. 
A reference image undergoes a random number ($k$) of noise-adding steps and then denoising with text guidance.
Finally, low-confidence detections are filtered out. 
}
\label{fig:ppl}
\vspace{-8pt}
\end{figure*}
\noindent\textbf{The {semantic} information in LDM} could be investigated through the cross-attention between the visual and textual information in the text-conditioned image generation process. 
We first compute the average cross-attention maps across all the time steps for each object in the text prompt, then binarize them with the OTSU's auto-thresholding method~\cite{otsu1979threshold}.
Results in Fig.~\ref{fig:feature}(d) demonstrate that the binarized cross-attention maps highlight the related object regions, indicating that the semantic guidance provided in the text prompt is well-reflected in LDM features.

\noindent\textbf{Is Cross-Attention Sufficient for Detection?}
\label{sec:compare_ca}
Binarizing the cross-attention (BCA) seems like a direct and explainable approach to generating bounding boxes for object detection.
Figs.~\ref{fig:feature}(a)(b) compare the detection results of our DiffusionEngine to the BCA.
As shown in Fig.~\ref{fig:feature}, the BCA suffers from four important limitations: 
i)~\underline{\textit{Instance Merging.}}  BCA reflects patch-wise text-visual similarity rather than instance understanding, leading to instance merging when there are overlapping instances of the same category (\emph{e.g.}, the two cats). 
ii)~\underline{\textit{Instance Chunking.}} BCA chunks an instance into multiple parts when partially obscured by other objects (\emph{e.g.}, the couch). 
iii)~\underline{\textit{Forcing Detect.}} BCA produces interpretable results only when the object is actually present in the image. When an object in the text prompt fails to be generated, BCA detection produces unexpected noise results (\emph{e.g.}, the remote). 
iv)~\underline{\textit{Missing Detect.}}, there could be unexpected objects in the image that are not present in the text prompt, causing missing detections (\emph{e.g.}, the potted plant is detected by DE (a) but missing in BCA (b)). 

To address these issues, we propose to learn a detection adapter that aligns the semantic and location information in LDM with detection-aware signals for improved detection.

\subsection{Learning Detection-Adapter}
\label{sec:train}

\noindent \textbf{Adapter Architecture.}
As shown at the top of Fig.~\ref{fig:ppl}, the proposed DiffusionEngine is comprised of a frozen diffusion model and a detection adapter that is designed to produce accurate detection bounding boxes. 
It is worth noting that any detection framework can be employed as the detection adapter.
We use the state-of-the-art detection framework DINO~\cite{zhang2022dino} in this paper. 
Specifically, the feature maps are extracted from each U-Net block, and groups of feature maps with the same resolution are concatenated to form a pyramid.
The detection adapter then utilizes the pyramid feature for predicting the bounding box $\hat y$.

\noindent \textbf{Adapter Optimization.}
In order to optimize the detection adapter, we require pairs of aligned LDM features and ground-truth detection results. While a naive approach to collect such data would be to generate a new synthetic dataset and extract features during the image generation process, this method is impractical due to the lack of ground-truth detection results and the burden of annotation. 
To circumvent this issue, we propose to leverage existing object detection benchmarks for adapter learning, where the LDM features are obtained by simulating the last denoising step with real images. Our training procedure is illustrated at the top of Fig.~\ref{fig:ppl}. Given an image $x\in\mathbb{R}^{H\times W\times 3}$ with its ground-truth annotations $y$, the encoder $\mathcal{E}$ first encodes $x$ into the latent representation $z_0=\mathcal{E}(x)$, where $z_0\in\mathbb{R}^{h\times w\times c}$ with $h=H/8$ and $w=W/8$. Then, a single forward diffusion step is performed to obtain the penultimate noisy sample $z_1$ using Eq.~\ref{eq:forward_diffusion} for $t=1$. Finally, by feeding $z_1$ and the time step $t$ into the U-Net for one-step denoising, we can extract intermediate features that approximate the features from the last image generation step:
\begin{equation}
\label{eq:one-step-back}
    \hat{z_0} = \epsilon_\theta(z_1, 1, c_{\emptyset}),
\end{equation}
where $c_{\emptyset}=\tau_\theta(\emptyset)$ is equivalent to the unconditional signal.
The chosen detection framework determines the training objective and can be simplified to:
\begin{equation}
    \mathcal{L}_{DE}=\mathcal{L}_{Det}(y, \hat y).
\end{equation}
We have empirically found that whether or not using an image-aligned text prompt during training has little effect on the training procedure ($c_p$ in Eq.~\ref{eq:backward_diffusion} versus $c_{\emptyset}$ in Eq.~\ref{eq:one-step-back}), as the conditioning signal has a negligible impact on the generation results in only one step of denoising. 

\noindent \textbf{Discussions.}
This one-step training procedure offers three main advantages: 
i) It only requires image-detection pairs for training, which allows for the use of datasets without corresponding image descriptions; 
ii) The layout and components of the original image are well-preserved after the inversion, which ensures the credibility of ground-truth annotations. 
iii) Existing labeled detection benchmarks can be directly used to learn the detection adapter, without additional data collection and labeling efforts.

\subsection{Scaling Up Data with DiffusionEngine}
\label{sec:generation}

Our DiffusionEngine, equipped with the learned detection adapter, can effectively scale up data in a single stage. 

\noindent \textbf{Image.}
By learning a detection adapter without modifying the LDM,  the image generation process remains identical to the original process in LDM. 
As shown at the bottom of Fig.~\ref{fig:ppl}, the reference image is first encoded and forwarded til a random noise-adding step $k$ using Eq.~\ref{eq:forward_diffusion}.
Then, the noisy sample $z_k$ is denoised for $k$ steps to generate the image guided by the text embedding. 
Since we no longer need to maintain the layout as in the training stage, we can fully utilize all image-generation capabilities inherited from LDM.

\noindent \textbf{Label.}
Consistent with the training process, we extract features from the last denoising step, feed them to the adapter, and obtain detection results for the generated image. Following the empirical practice in detector inference, we filter out low-confidence predictions with a threshold $\delta=0.3$, keeping the rest as generated annotations.

\noindent \textbf{Diversity.}
By modifying the seed, encode ratio, guidance scale, and conditional text prompt, our DE can scale up the reference dataset with labeled generated images that have various degrees of discrepancy to the reference images. The second row in Fig.\ref{fig:head} provides an example of data scaling-up using different encode ratios. As the noise-adding step increases, slight distortion accumulates, resulting in more diverse reconstructed images compared to the original input. Our DE generates labeled data well for multi-object tasks with different sizes and is not limited to the original layout.

\noindent \textbf{Prompts.}
For datasets that have off-the-shelf captions for each image, we directly use these captions as input text prompts for image generation. For those without captions, we use a generic text prompt, {'A [domain], with [cls-a], [cls-b], ... in the [domain].'}, where {[cls-i]} represents the object names appearing in each image and the [domain] tag is curated respect to the data, \emph{e.g.}, photo, clipart.

\section{DiffusionEngine Detection Dataset}
In this section, we detail the construction of our two scaling-up datasets, termed COCO-DE and VOC-DE. 
The statistics of the two datasets are summarized in Tab.~\ref{tab:de-dataset}.

\noindent\textbf{Reference Images \& Text Prompts.}
We employ an image-guided text-to-image generation process to scaling-up datasets.
For COCO-DE, we adopt the images from COCO train2017 as references and their corresponding captions as text prompts.
For VOC-DE, the images are from the Pascal VOC trainval0712 split, and we use the generic text prompt as described in the former section.

\noindent\textbf{Image Size.}
The reference image is resized to $512\times1024$ with the original aspect ratio then random crop within 768$^2$.

\noindent\textbf{Image Diversity.}
For each image, we randomly choose a seed and an encoding ratio between 0.3 and 1.0 to ensure generative diversity. When the encoding ratio is set to 1.0, the image is converted to Gaussian noise, and the generation process is collapsed to the text-to-image generation process.

\noindent\textbf{Annotation Diversity.}
We establish an annotation lower bound and record the number of generated annotations for each category during the scaling-up procedure. This process ends when all categories exceed the target lower bound.

\begin{table}[th]
\centering
\footnotesize
\caption{Dataset Statistics of COCO-DE and VOC-DE.}
\label{tab:de-dataset}
\begin{tabular}{l|rr|rr}
\toprule
  Dataset &  \#Images & \#Scale &  \#Instances & \#Scale \\
\midrule
COCO & 117,266 & - & 849,949  & - \\
COCO-DE & 205,287 & \textbf{1.7$\times$} & 1,281,418 & \textbf{1.5$\times$} \\
\midrule
VOC & 16,551 & - & 47,223 & - \\
VOC-DE & 64,934 & \textbf{3.9$\times$} & 168,141 & \textbf{3.6$\times$} \\
\bottomrule
\end{tabular}
\end{table}

\section{Experiments}

\subsection{Implementation Details}
We freeze the pre-trained Stable Diffusion v2 and optimize the detection-adapter solely on COCO~\cite{coco} without additional data.
The adapter is trained for 90k iterations with a global batch size of 64.
AdamW~\cite{adamw} is employed, with the \textit{lr} starting at 2e-4 and decreases to 2e-5 at the 80k iteration. 
For data scaling-up, we use DPM-Solver++ as the sampler, with default inference steps of 30 and a classifier-free guidance scale of 7.5.

\subsection{COCO Detection Evaluation}
We evaluate the effectiveness of scaling up data using DiffusionEngine (DE) on the widely-used COCO object detection benchmark. 
\textit{For fair comparison, we maintain identical batch sizes, apply the same data augmentations, and conduct an equal number of training iterations in each experiment group.} 
We used the default settings for each algorithm as in \cite{mmdetection, detectron2}.
To simplify the exposition, we utilize ``DE'' to denote data scaling-up via DiffusionEngine in subsequent sections. The outcomes are detailed in Table~\ref{tab:orth}, which indicates that DE is orthogonal to existing works in the following aspects:

\noindent\textbf{Detection Algorithms.}
We adopt various detection algorithms, including the anchor-based one-stage algorithm \textit{RetinaNet}~\cite{retinanet}, anchor-based two-stage algorithm \textit{Faster-RCNN}~\cite{ren2015faster}, and anchor-free algorithm \textit{DINO}~\cite{zhang2022dino}.
The results show that incorporating data generated via DE outperforms the baseline by e.g. 3.3\% and 3.1\% mAP with RetinaNet and DINO (ResNet50), respectively. This demonstrates that combining DE-generated data with different detection algorithms achieves consistent performance gains.

\begin{table*}[t]
\footnotesize
\centering
\caption{Effectiveness of DE on {COCO}. We combine the DE-generated training data with different \textbf{Frameworks, Backbones, Pre-train Dataset}. Consistent improvement demonstrates the proposed DE is effective and orthogonal to existing methods. 
}
\label{tab:orth}
\resizebox{%
      \textwidth
    }{!}{%
\begin{tabular}{l|c|c|c|llllll}
\toprule
Framework  & Backbone  & Pre-train & Schedule  & mAP  & mAP$_{50}$ & mAP$_{75}$ & mAP$_s$  & mAP$_m$ & mAP$_l$ \\ \midrule
RetinaNet  &  &  &  & 38.0 & 57.0 & 40.6 & 22.2 & 41.2 & 49.3  \\
\rowcolor[HTML]{EFEFEF} 
w/ DE  & \multirow{-2}{*}{R50} & \multirow{-2}{*}{IN-1k} & \multirow{-2}{*}{6$\times$} & 41.3 \gain{3.3}  & 60.4 \gain{3.4} & 44.2 \gain{3.6} & 24.1 \gain{1.9} & 45.9 \gain{4.7} & 53.8 \gain{4.5} \\ 
\midrule
Faster-RCNN  &  &  &  & 39.0 & 59.4 & 42.4 & 23.2 & 41.7 & 51.3 \\
w/ BYOL~\cite{byol} &  &  &  & 40.4 & 60.6 & 44.0 & 24.4 & 43.5 & 51.8   \\
\rowcolor[HTML]{EFEFEF} 
w/ DE + BYOL   & \multirow{-3}{*}{R50}  &  \multirow{-3}{*}{IN-1k} &  \multirow{-3}{*}{9$\times$} & 43.8 \gain{4.8} & 63.7 \gain{4.3} & 47.4 \gain{5.0}   & 25.8 \gain{2.6} & 47.8 \gain{6.1} & 57.3 \gain{6.0} \\ \midrule
DINO    &  &  &  & 49.2  & 67.0  & 53.6 & 32.9 & 52.3 & 62.8  \\
\rowcolor[HTML]{EFEFEF}
w/ DE  & \multirow{-2}{*}{R50} & \multirow{-2}{*}{IN-1k} & \multirow{-2}{*}{6$\times$} & 52.3 \gain{3.1}  & 70.0 \gain{3.0} & 57.1 \gain{3.5} & 35.3 \gain{2.4} & 56.0 \gain{3.7} & 66.3 \gain{3.5}  \\  \midrule
\addlinespace[3pt] DINO   &    &   &   & 57.1 & 76.1   & 62.2 & 39.0 & 61.3 & 72.9   \\
\rowcolor[HTML]{EFEFEF}w/ DE   &  \multirow{-2}{*}{Swin-L}   &  \multirow{-2}{*}{IN-22k} &  \multirow{-2}{*}{9$\times$}     & 58.8 \gain{1.7}    & 77.1 \gain{1.0}   & 64.3 \gain{2.1} & 41.8 \gain{2.8} & 62.6 \gain{1.3} & 75.0 \gain{2.1}   \\ \bottomrule
\end{tabular}%
}
\end{table*}

\begin{figure}[ht]
	\centering
	\includegraphics[width=\linewidth]{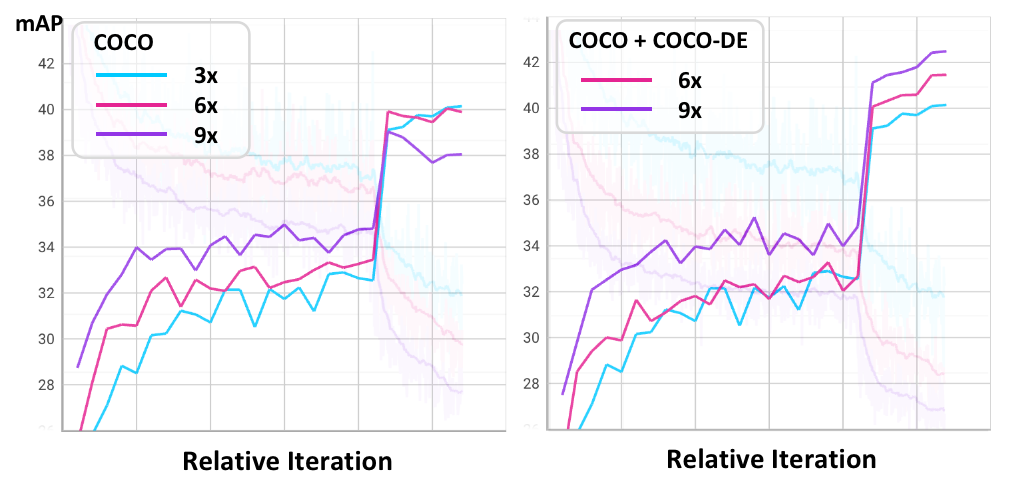}
	\caption{Performance with increasing schedule on COCO v.s COCO w/DE.}
 	\label{fig:ablation_schedule}
  \vspace{-10pt}
\end{figure}

\noindent\textbf{Backbone Pre-training Algorithm.} In addition to {fully-supervised} pre-training, we use {self-supervised} pre-training backbone~\cite{byol} as initialization to validate the robustness of DE. The second block of Tab.~\ref{tab:orth} shows that it leads to a 1.4\% mAP improvement over the fully-supervised baseline. Moreover, combining DE with the self-supervised pre-trained backbone further boosts mAP to 43.8\%, which is an additional gain of 3.4\% compared to the previous result.

\noindent\textbf{Backbone and Pre-training Dataset.} In the last two blocks, we conduct experiments on the DINO framework with two backbones: ResNet50~\cite{he2016resnet} and Swin-L~\cite{liu2022swin} Transformer. 
DE provides a 3.1\% mAP gain with the ResNet50 backbone and +1.7\% with the Swin-L. 
Even starting with a strong baseline with large backbone architecture (Swin-L) and pre-trained on a bigger dataset (ImageNet-22k), DE can further boost the performance.

\noindent\textbf{Performance with Schedule Scaling.}
Fig.~\ref{fig:ablation_schedule} depicts the performance curves for training with or without the generated COCO-DE for various schedules. 
The X-axis displays the relative iterations w.r.t. the declined learning rate.
Here we have three observations:
i) validation performance using the 3$\times$ schedule gradually improves as expected.
ii) without generated data, increasing schedule leads to a decrease in both validation mAP and the training loss, \emph{i.e.}, overfitting occurs.
iii) scaling data with DE further improves the performance, indicating that DE  is an effective data scaling-up technology rather than a simple data replay.
We observe the same tendency of overfitting for all baselines in Tab. \ref{tab:orth}, even for DINO with strong default data augmentation (see supp.).

\noindent\textbf{Generalization.} 
We also experiment on the VOC-0712 dataset (Tab.~\ref{tab:voc0712})
to verify the generalization of DiffusionEngine.
When training only with the generated data, we could already surpass the baseline that training on the real, manually labeled dataset.
By combining the generated dataset with the real labeled data, we achieve further improvement, which indicates that the DE-generated data is an effective supplement to the real dataset. 

\begin{table}[hb]
\centering
\footnotesize
\caption{DE on {VOC-0712}. The backbone is ResNet50. $^\dagger$ indicates annotations of real images are not used for training.}
\label{tab:voc0712}
\resizebox{\columnwidth}{!}{
\begin{tabular}{l|c|l|c|c}
\toprule 
Method & \multicolumn{1}{c|}{\#Images} & mAP  & mAP$_{50}$  & mAP$_{75}$ \\ \midrule
 Faster-RCNN & 16551 (1$\times$) & 50.7 & 80.2 & 55.0 \\
\rowcolor[HTML]{EFEFEF}\cellcolor[HTML]{EFEFEF} & 5$\times^\dagger$  & 52.5 \gain{1.8} & 77.2  & 58.1 \\ 
 \rowcolor[HTML]{EFEFEF}\multirow{-2}{*}{\cellcolor[HTML]{EFEFEF}w/ DE} & 6$\times$  & 58.3 \gain{7.6} & 82.7 & 64.7  \\
\bottomrule
\end{tabular}
}%
\end{table}

\subsection{Comparison with SOTAs}
This section compares DE with some state-of-the-art data scaling-up techniques, such as Copy-Paste~\cite{ghiasi2021simple} and DALL-E for Detection~\cite{ge2022dall}. Following \cite{ge2022dall}, we use Faster-RCNN with ResNet-50 as the backbone and experiment on the VOC2012 segmentation set. As shown in Tab.~\ref{tab:cmp-dalle}, the relative gain of DE surpasses that of DALL-E by adding only twice the amount of original data, even surpassing the strong baseline Copy-Paste~\cite{ghiasi2021simple}, demonstrating that DE helps provide higher-quality pairs. 
Although copy-paste infinitely scales up the amount of training data through random combination, its diversity is limited by the original instances, while DE does not. 
We can see that the performance continues to increase as more generated data is added, indicating that DE is an effective solution for large-scale data expansion. 
We also compare with X-Paste~\cite{zhao2022x} under the same  setting that the baseline CenterNet2~\cite{zhou2021probabilistic} is trained with the Swin-L backbone on COCO.
Results show that the relative mAP improvement by the proposed DE without using the mask for training is 2.0\%, which is higher than 1.5\% improvement achieved by X-Paste. 

\begin{table}[htb]
\centering
\footnotesize
\caption{Compare with SOTAs on {VOC-12}. We only use the Segmentation Set for experiments following DALL-E, but the seg. masks are \textbf{NOT} used in our experiments. $^*$ denotes our reproduced result in the same setting.}
\label{tab:cmp-dalle}
\resizebox{\columnwidth}{!}{
\begin{tabular}{l|c|l|c|c}
\toprule 
Method & \multicolumn{1}{c|}{\#Images} & mAP  & mAP$_{50}$  & mAP$_{75}$ \\ \midrule
Faster-RCNN               & \multicolumn{1}{c|}{1464 (1$\times$)}    & 17.0   & 45.5  & - \\ 
w/ DALL-E   & 41$\times$  & 25.9 {\textcolor{gray}{(8.9{$\uparrow$})}} & 51.8  & - \\ 
\midrule
Faster-RCNN$^*$              & \multicolumn{1}{c|}{1464 (1$\times$)}                          & 18.1 & 44.9 & 9.8 \\ 
 w/ Copy-Paste & - & 24.5 {\textcolor{gray}{(6.5{$\uparrow$})}} & 54.9 & 17.2 \\
\rowcolor[HTML]{EFEFEF}\cellcolor[HTML]{EFEFEF} & 2$\times$              & 26.1 \gain{8.1} & 57.9 & 18.9   \\ 
\rowcolor[HTML]{EFEFEF}\cellcolor[HTML]{EFEFEF} & 3$\times$              & 30.0 \gain{11.9} & 63.1 & 23.3   \\
\rowcolor[HTML]{EFEFEF}\cellcolor[HTML]{EFEFEF} & 5$\times$              & 34.2 \gain{16.1} & 67.4 & 29.4   \\
 \rowcolor[HTML]{EFEFEF}\multirow{-4}{*}{\cellcolor[HTML]{EFEFEF}w/ DE} &  9$\times$  & 39.0 \gain{20.9} & 71.6 & 37.9  \\
\bottomrule
\end{tabular}
}%
\end{table}

\begin{figure*}[t]
	\centering
	\includegraphics[width=\linewidth]{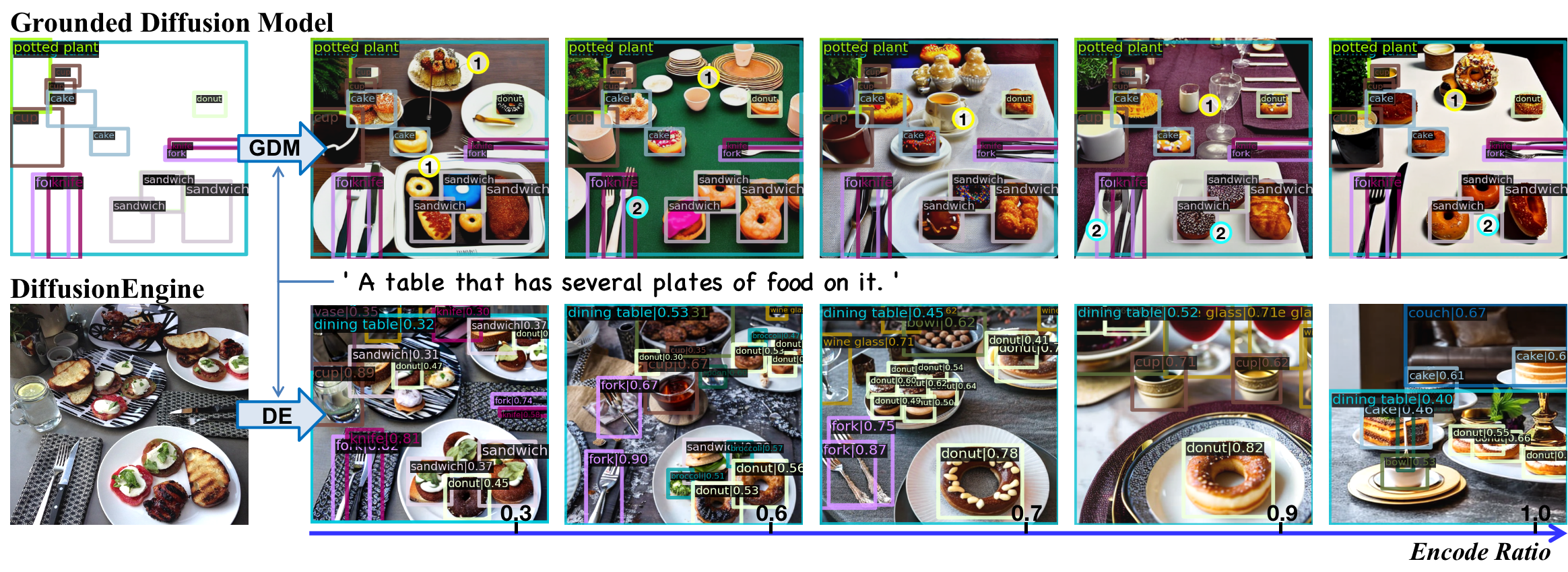}
	\caption{
Comparison with Grounded Diffusion Model (GDM). Scaling up data with GLIGEN~\cite{li2023gligen} and our DE follows a distinct paradigm. While GDM specifies the layout and explicitly controls the image generation, our DE predicts the layout concurrently with the generation process. As depicted, GDM may generate unexpected objects in unspecified areas, leading to missed annotation\ding{172}, whereas the annotations of specified areas may be wrong due to mistaken generation\ding{173}.}
 	\label{fig:cmp_grounded}
  \vspace{-10pt}
\end{figure*}

\subsection{Cross Domain Data Scaling-up}
To assess the robustness of DiffusionEngine in out-of-domain scenarios, we conducted experiments on the Clipart-1k dataset~\cite{clipart}, which comprises 500 Clipart domain images.
As indicated in the first block of Tab.~\ref{tab:da}, we directly trained the model using the DE-generated dataset. The results indicate that DE significantly outperforms the model trained on Clipart (+11.5\%), thus highlighting the efficacy of DiffusionEngine in scaling up cross-domain data.
In addition, we leveraged Adaptive Teacher (AT~\cite{li2022cross}) to perform cross-domain semi-supervised experiments. As demonstrated in the second block of Tab.~\ref{tab:da}, incorporating DE-generated images as either unlabeled or labeled data yielded gains of 7.9\% and 14.1\%, respectively. These findings validate the robustness of DE to generate images and labels for semi-supervised learning.

\begin{table}[t]
\centering
\caption{The results of cross-domain object detection on the Clipart1k test set for {VOC-12 → Clipart-1k} adaptation. $^\dagger$ indicates annotations of real images are not used for training.
}
\label{tab:da}
\footnotesize
\resizebox{%
      \linewidth
    }{!}{%
\begin{tabular}{c|c|c|l}
\toprule
 & {Labeled}    & {Unlabeled} & mAP$_{50}$ \\ 
\midrule
   & {VOC} & - & 28.8 \\
   & {Clipart} & - & 45.0 \\ 
\rowcolor[HTML]{EFEFEF} \multirow{-3}{*}{\cellcolor{white} Sup}  & {DE$^\dagger$}  &- & 56.5 \gain{11.5} \\
 \midrule
    & {VOC}   & Clipart & 49.3  \\
\rowcolor[HTML]{EFEFEF}\cellcolor{white} & VOC      & Clipart + DE$^\dagger$              & 52.9 \gain{7.9}               \\ 
\rowcolor[HTML]{EFEFEF} \multirow{-3}{*}{\cellcolor{white}\shortstack[c]{Semi-Sup \\(AT~\cite{li2022cross})}}  &  VOC+DE$^\dagger$ &  Clipart & 63.4 \gain{14.1} \\
\bottomrule
\end{tabular}%
}
\vspace{-5pt}
\end{table}

\subsection{Discussion with Grounded Diffusion Models}
We also investigate recent grounded diffusion models (GDMs) such as ReCo~\cite{yang2022reco} and GLIGEN~\cite{li2023gligen}, and compare with DiffusionEngine:

\noindent\textbf{\textit{Paradigm}}: GDMs are primarily designed to generate controllable results based on detection boxes, whereas DiffusionEngine strives to generate diverse images with accurate annotations via a single-step inference.

\noindent\textbf{\textit{Condition}}: GDMs necessitate category lists, prompts, and additional bounding boxes, DiffusionEngine only require simple text prompts and optional reference images.

\noindent\textbf{\textit{Performance}}: 
As shown in Fig.~\ref{fig:cmp_grounded}, DiffusionEngine effectively unifies the processes of image generation and labeling, thereby enabling the provision of a wide variety of images with detailed annotations. In contrast, GDM is limited by the conditions of the box and leads to missed annotations, mistaken image generation, and simplistic layouts.

\section{Limitations and Further Work}

\noindent\textbf{All-in-One Model.} DiffusionEngine can be easily extended to other tasks via task-specific adaptors. 

\noindent\textbf{ChatGPT.} Textual guidance prompts are not available in many scenarios. It would be interesting to introduce ChatGPT with in-context learning to generate guidance prompts.

\noindent\textbf{RLHF.}  
Integrating task-aware human feedback may further improve the alignment and quality of detection pairs.

We hope that our work will inspire more researchers to investigate data scaling-up using the diffusion model and provide valuable insights for future research.

\section{Conclusion}
We introduce the DiffusionEngine (DE), a scalable and efficient data engine for object detection that generates high-quality detection-oriented training pairs in a single stage. 
The detection-adapter aligns the implicit detection-oriented knowledge in off-the-shelf diffusion models to generate accurate annotations. 
Additionally, we contribute two datasets, COCO-DE and VOC-DE, which are intended to scale up existing detection benchmarks. 
Our experiments demonstrate that DE enables the generation of scalable, diverse, and generalizable data, and incorporating data scaling up via DE through a plug-and-play manner can achieve significant improvements in various scenarios.

{\small
\bibliography{DEDbib}
}
\newpage
\twocolumn[\centering\textbf{\LARGE Supplementary}\bigskip]

\setcounter{section}{0}

\section{Implementation Details}
\subsection{Prompts for Scaling-up Data}
We use the following additional prompts for improving \underline{\textit{photo}} generation quality:

\noindent\textbf{Positive Prompt:}
elegant, meticulous, magnificent, maximum details, extremely hyper aesthetic, highly detailed.

\noindent\textbf{Negative Prompt:}
naked, deformed, bad anatomy, out of focus, disfigured, bad image, poorly drawn face, mutation, mutated, extra limb, ugly, disgusting, poorly drawn hands, missing limb, floating limbs, disconnected limbs, blurry, mutated hands and fingers, watermark, oversaturated, distorted hands.

These prompts tend to generate realistic photos, so they are not used when generating \underline{\textit{clipart}}.

\subsection{Training Schedule}
Following the common experiment setup, we refer 90k iterations with batch size 16 to a "1$\times$ schedule", and the final number of schedules is based on the total learning samples.
Note that the only difference between ours and the baseline is the addition of annotated training data produced by the proposed DiffusionEngine (DE), while maintaining equal total iterations. 
The training setup is detailed as follows:

\begin{table}[th]
\centering
\footnotesize
\caption{Details for Schedule in Table 2 (manuscript).}
\resizebox{%
      \linewidth
    }{!}{%
\begin{tabular}{l|ccc|c}
\toprule
  Model & Batchsize & LR & Total Iter. & Schedule \\
\midrule
RetinaNet-R50 & 32 & 0.02 & 270k & 6$\times$ \\
Faster-RCNN-R50 & 48 & 0.04 & 270k & 9$\times$ \\
DINO-R50 & 32 & 2e-4 & 270k & 6$\times$ \\
DINO-Swin-L & 48 & 2e-4 & 270k & 9$\times$ \\
\bottomrule
\end{tabular}
\label{tab:schedule_detail}
}%
\end{table}

\section{Performance with Schedule Scaling}
\begin{figure}[hb]
	\centering
	\includegraphics[width=.8\linewidth]{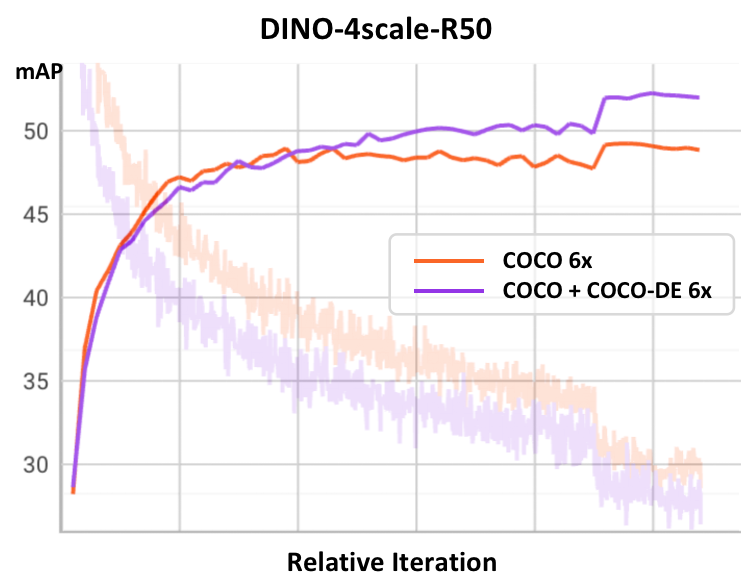}
	\caption{Performance with increasing schedule on COCO v.s COCO w/DE.}
 	\label{fig:ablation_schedule_dino}
\end{figure}

As shown in Figure \ref{fig:ablation_schedule_dino}, the tendency of overfitting also occurred for DINO with strong default data augmentation, while combining COCO-DE alleviates the issue.

\section{Performance Gain Analysis}
\begin{figure}[htb]
	\centering
	\includegraphics[width=\linewidth]{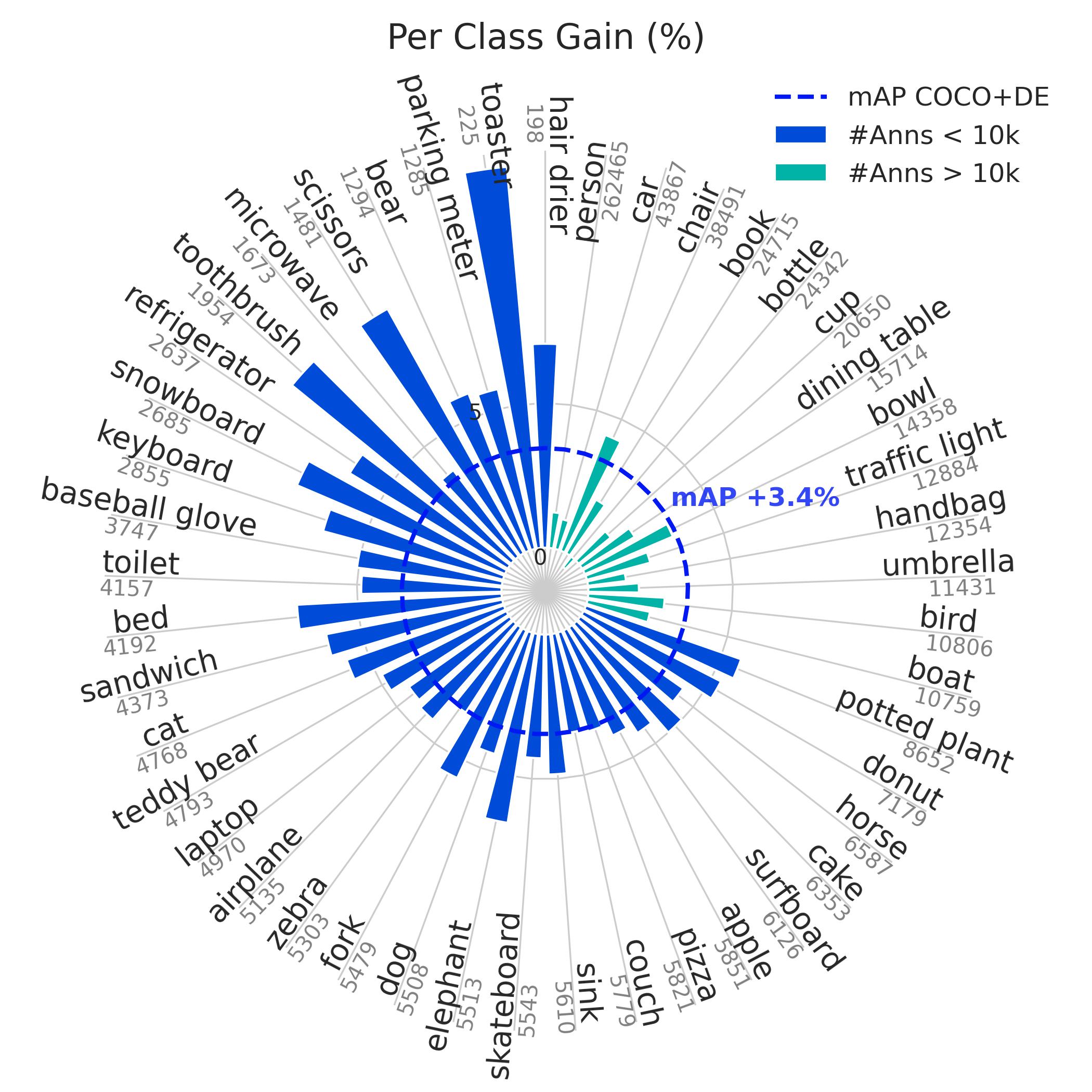}
	\caption{Analyze Performance Gain by Category.}
 	\label{fig:coco100_per_cls}
\end{figure}

To investigate the performance gain of DE, we further analyze the improvement in category based on Faster-RCNN. 
In Figure~\ref{fig:coco100_per_cls}, we sort the categories according to the number of annotations in COCO.
Each bar represents the precision gain over the baseline for a specific category.
It can be observed that the mAP gain mainly comes from classes with fewer annotations in the original dataset (less than 10k), indicating that DE helps to alleviate the lack-of-sample problem.

\section{More Qualitative Results}
\begin{figure*}[th]
\centering
\includegraphics[width=.95\linewidth]{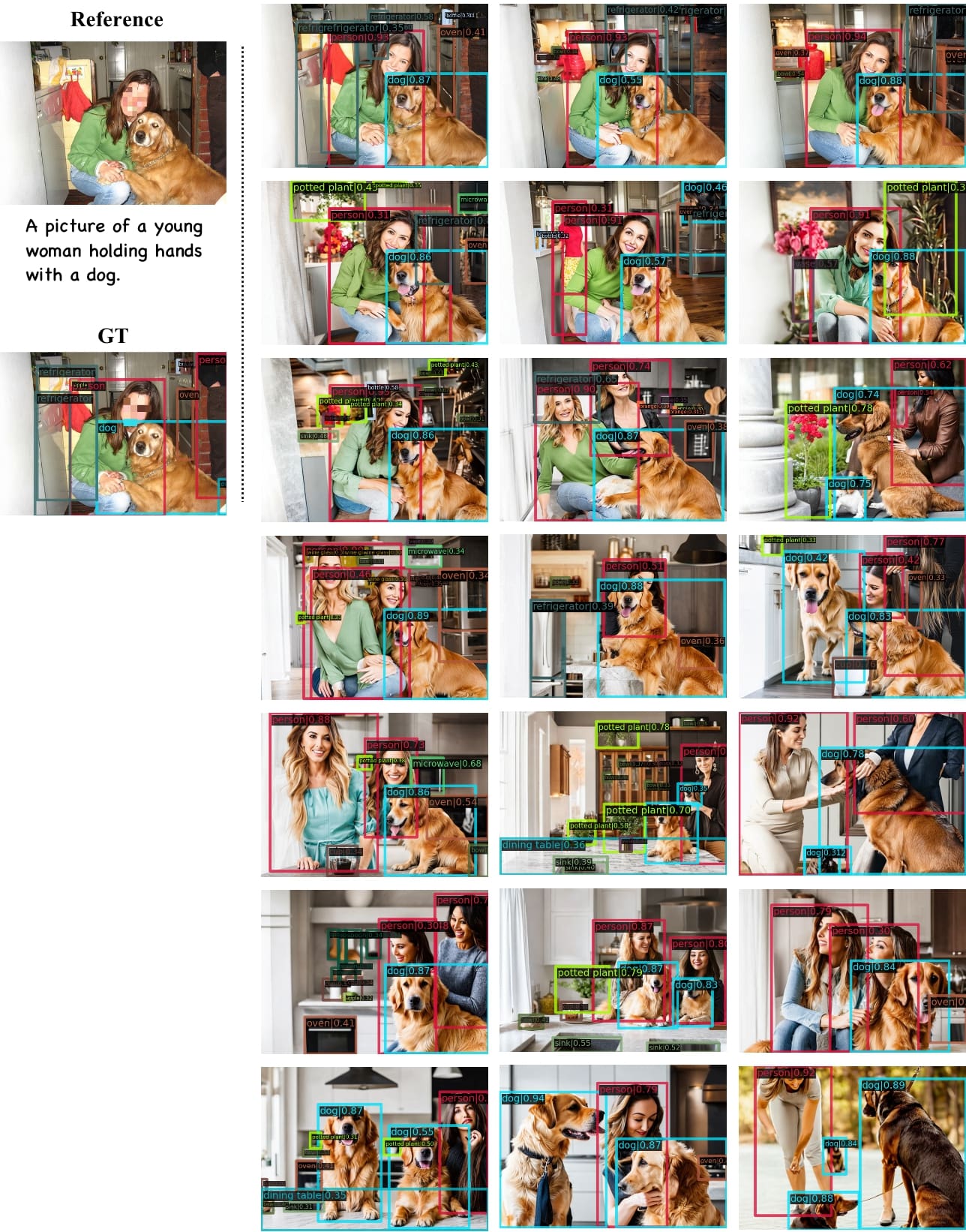}
\caption{Visualization on Scalable and Diverse generation of photo with DiffusionEngine.
}
\label{fig:s_photo_1}
\end{figure*}

\begin{figure*}[th]
\centering
\includegraphics[width=.9\linewidth]{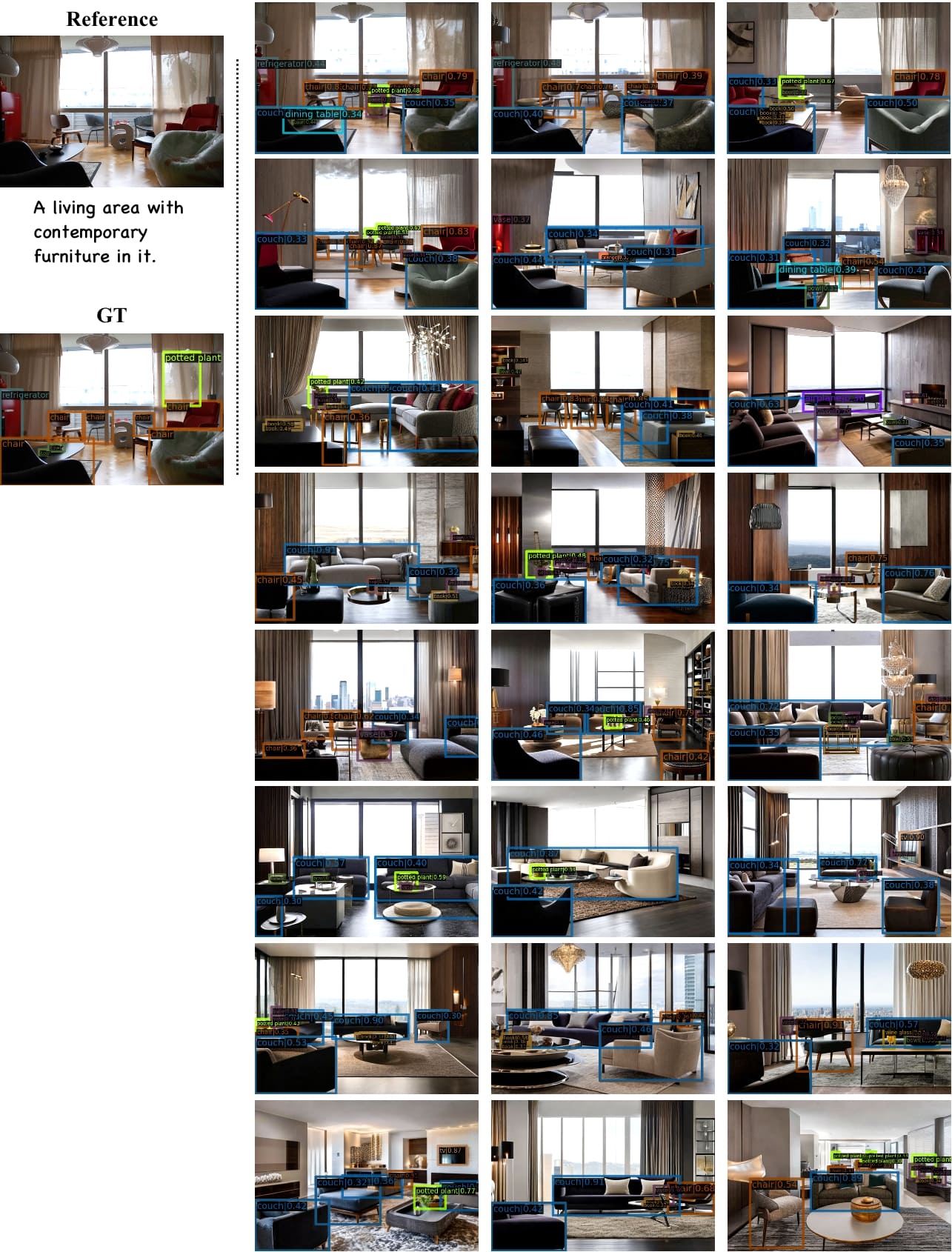}
\caption{Visualization on Scalable and Diverse generation of photo with DiffusionEngine.
}
\label{fig:s_photo_2}
\end{figure*}

\begin{figure*}[th]
\centering
\includegraphics[width=.95\linewidth]{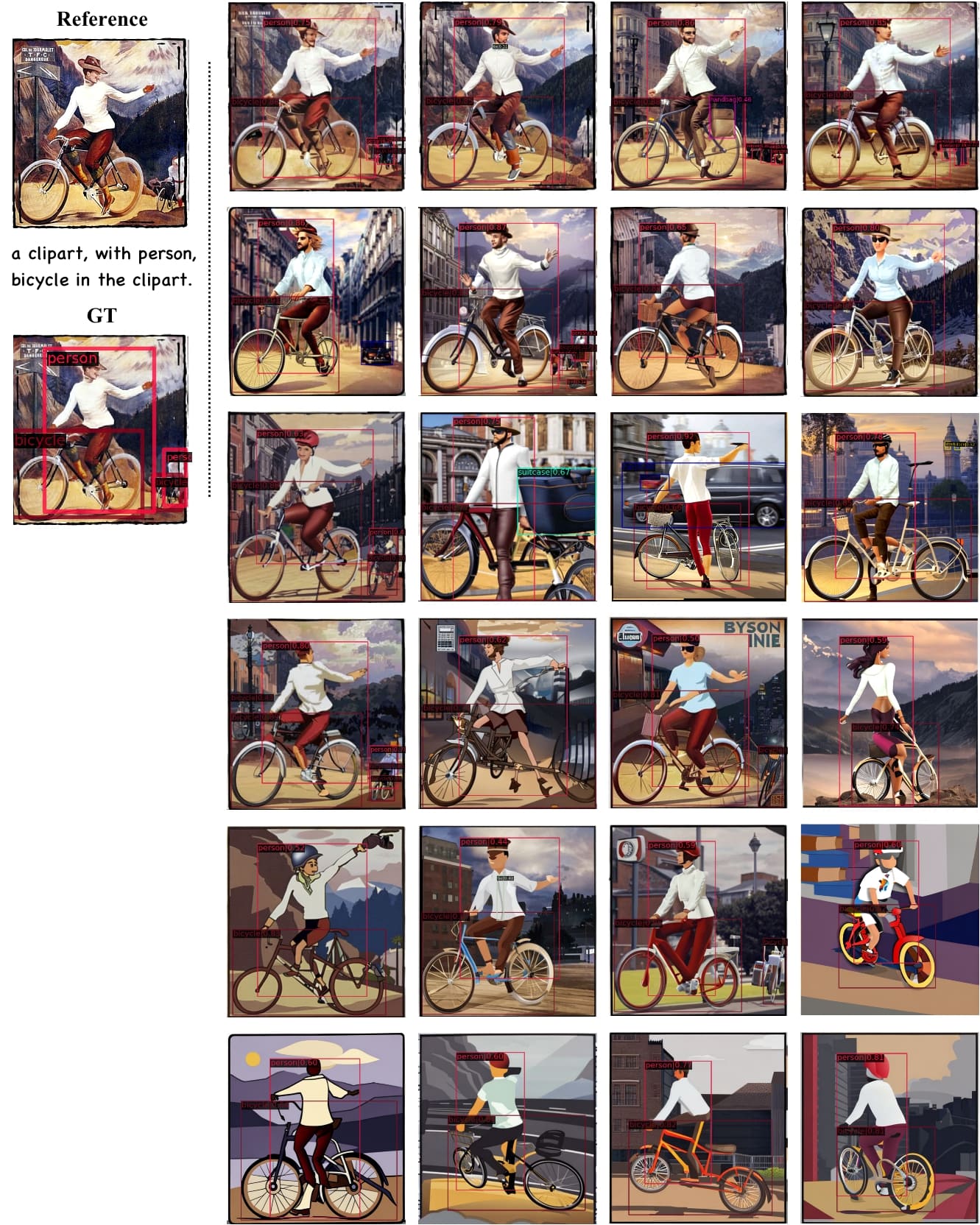}
\caption{Visualization on Scalable and Diverse generation of clipart with DiffusionEngine.
}
\label{fig:s_clipart_1}
\end{figure*}

\begin{figure*}[th]
\centering
\includegraphics[width=\linewidth]{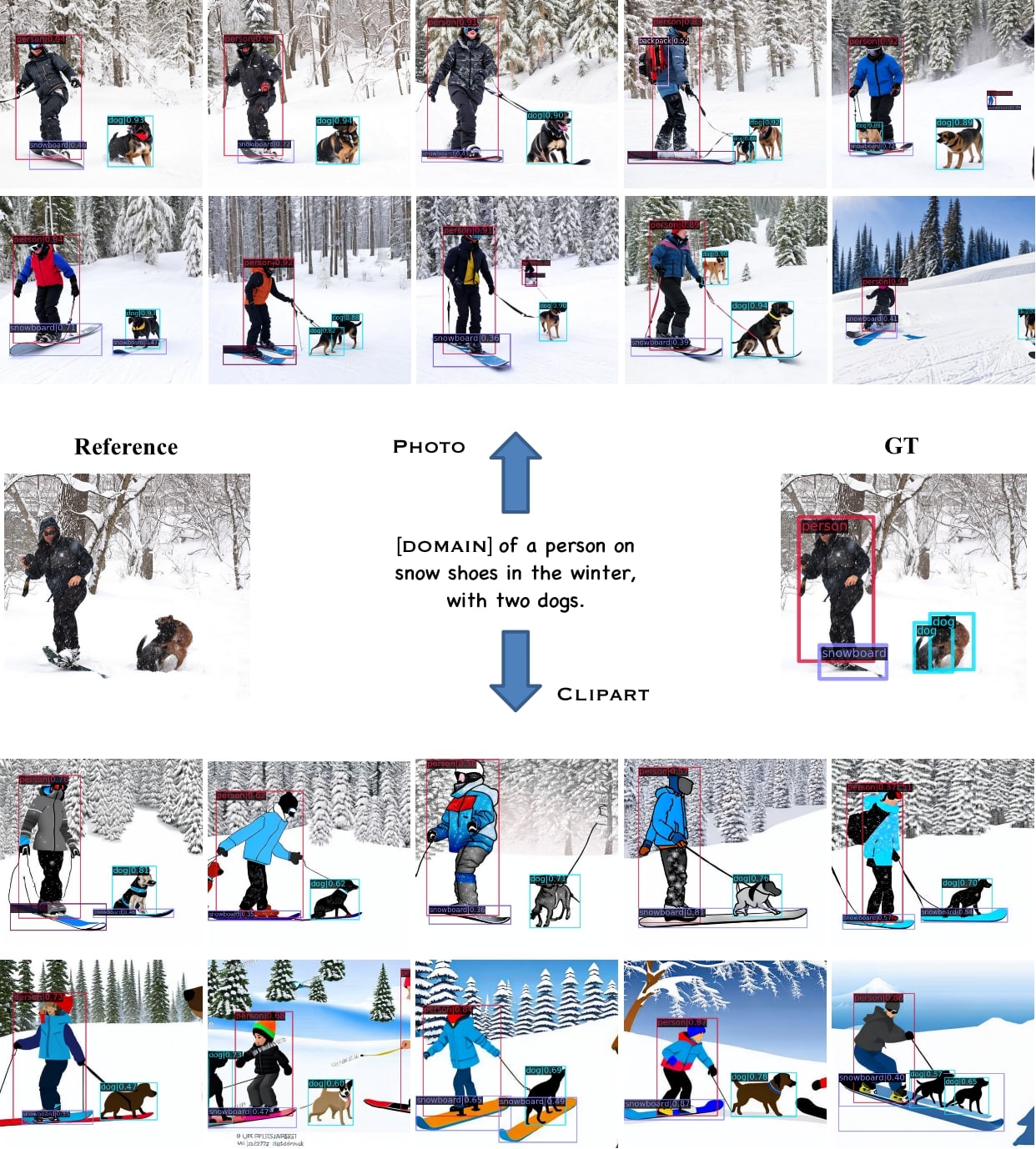}
\caption{Visualization on generalization ability of DiffusionEngine.}
\label{fig:s_cross_1}
\end{figure*}

Here we provide more visualization results of data scaling up for photo (Figure~\ref{fig:s_photo_1}, \ref{fig:s_photo_2}), and clipart (Figure~\ref{fig:s_clipart_1}).
We also show that DiffusionEngine generalizes well across domains by simply modifying the prompt (Figure~\ref{fig:s_cross_1}). 
The ground-truth (GT) annotations for the reference images are shown but not used in our generation process.


\end{document}